\documentclass[11pt,a4paper]{article}
\usepackage[hyperref]{naaclhlt2018}
\usepackage{times}
\usepackage{latexsym}
\usepackage{graphicx}
\usepackage{multirow}
\usepackage{amssymb,amsmath,bm}
\usepackage{algorithm}
\usepackage{algpseudocode}
\usepackage{booktabs}

\usepackage{url}

\DeclareMathOperator*{\argmax}{argmax\,}
\aclfinalcopy 

\title{Feudal Reinforcement Learning for Dialogue Management in Large Domains}

\author{I\~nigo~Casanueva$^{1}$, Pawe\l~Budzianowski$^1$, Pei-Hao~Su$^2$,\\
  \textbf{Stefan~Ultes$^1$, Lina~Rojas-Barahona$^1$, Bo-Hsiang Tseng$^1$ and Milica~Ga{\v s}i{\' c}$^1$}  \\
  $^1$Department of Engineering, University of Cambridge, UK \\
  $^2$PolyAI Limited, London, UK \\
  {\tt ic340@cam.ac.uk}
  \\
  }

\date{}

\begin{document}
\maketitle
\begin{abstract}
\renewcommand{\thefootnote}{\fnsymbol{footnote}} 
Reinforcement learning (RL) is a promising approach to solve dialogue policy optimisation. Traditional RL algorithms, however, fail to scale to large domains due to the curse of dimensionality. We propose a novel Dialogue Management architecture, based on Feudal RL, which decomposes the decision into two steps; a first step where a master policy selects a subset of primitive actions, and a second step where a primitive action is chosen from the selected subset. The structural information included in the domain ontology is used to abstract the dialogue state space, taking the decisions at each step using different parts of the abstracted state. This, combined with an information sharing mechanism between slots, increases the scalability to large domains. We show that an implementation of this approach, based on Deep-Q Networks, significantly outperforms previous state of the art in several dialogue domains and environments, without the need of any additional reward signal. 
\renewcommand{\thefootnote}{\fnnumerals{footnote}} 
\end{abstract}
\section{Introduction}
Task-oriented Spoken Dialogue Systems (SDS), in the form of personal assistants, have recently gained much attention in both academia and industry. One of the most important modules of a SDS is the Dialogue Manager (DM) (or policy), the module in charge of deciding the next action in each dialogue turn. Reinforcement Learning (RL) \citep{RL} has been studied for several years as a promising approach to model dialogue management \citep{levin1998using,henderson2008hybrid,pietquin2011sample,young2013pomdp,casanueva2015knowledge,su2016continuously}. However, as the dialogue state space increases, the number of possible trajectories needed to be explored grows exponentially, making traditional RL methods not scalable to large domains.

Hierarchical RL (HRL), in the form of temporal abstraction, has been proposed in order to mitigate this problem \citep{cuayahuitl2010evaluation,cuayahuitl2016deep,budzianowski2017subdomain,peng2017composite}. However, proposed HRL methods require that the task is defined in a hierarchical structure, which is usually handcrafted. In addition, they usually require additional rewards for each subtask. Space abstraction, instead, has been successfully applied to dialogue tasks such as Dialogue State Tracking (DST) \citep{Henderson2014b}, and policy transfer between domains \citep{gavsic2013pomdp,gavsic2015policy,wang2015learning}. For DST, a set of binary classifiers can be defined for each slot, with shared parameters, learning a general way to track slots. The policy transfer method presented in \citep{wang2015learning}, named Domain Independent Parametrisation (DIP), transforms the belief state into a slot-dependent fixed size representation using a handcrafted feature function. This idea could also be applied to large domains, since it can be used to learn a general way to act in any slot. 

In slot-filling dialogues, a HRL method that relies on space abstraction, such as Feudal RL (FRL) \citep{dayan1993feudal}, should allow RL scale to domains with a large number of slots. FRL divides a task spatially rather than temporally, decomposing the decisions in several steps and using different abstraction levels in each sub-decision. This framework is especially useful in RL tasks with large discrete action spaces, making it very attractive for large domain dialogue management. 

In this paper, we introduce a Feudal Dialogue Policy which decomposes the decision in each turn into two steps. In a first step, the policy decides if it takes a slot independent or slot dependent action. Then, the state of each slot sub-policy is abstracted to account for features related to that slot, and a primitive action is chosen from the previously selected subset. Our model does not require any modification of the reward function  and the hierarchical architecture is fully specified by the structured database representation of the system (i.e. the ontology), requiring no additional design. 


\section{Background}\label{sec:bgnd}
Dialogue management can be cast as a continuous MDP \citep{young2013pomdp} composed of a continuous multivariate belief state space $\mathcal{B}$, a finite set of actions $\mathcal{A}$ and a reward function $R(b_t,a_t)$. At a given time $t$, the agent observes the belief state $b_t \in \mathcal{B}$, executes an action $a_t\in \mathcal{A}$ and receives a reward $r_t \in \mathbb{R}$ drawn from $R(b_t,a_t)$. The action taken, $a$, is decided by the \textit{policy}, defined as the function $\pi(b)=a$. For any policy $\pi$ and $b \in \mathcal{B}$, the $Q$-value function can be defined as the expected (discounted) return $R$, starting from state $b$, taking action $a$, and then following policy $\pi$ until the end of the dialogue at time step $T$:
\begin{equation} \label{eq:value_fun}
Q^{\pi}(b,a) =  \mathbb{E}\{ R | b_t=b, a_t=a\} 
\end{equation}
where $R = \sum_{\tau=t}^{T-1} \gamma^{(\tau-t)} r_\tau$ and $\gamma$ is a discount factor, with $0 \leq \gamma \leq 1$.

The objective of RL is to find an optimal policy $\pi^{*}$, i.e. a policy that maximizes the expected return in each belief state. In \textit{Value-based} algorithms, the optimal policy can be found by greedily taking the action which maximises $Q^{\pi}(b,a)$.


In slot-filling SDSs 
the belief state space $\mathcal{B}$ is defined by the \textit{ontology}, a structured representation of a database of entities that the user can retrieve by talking to the system. Each entity has a set of properties, refereed to as \emph{slots} $\mathcal{S}$, where each of the slots can take a value from the set $\mathcal{V}_s$. The belief state $b$ is then defined as the concatenation of the probability distribution of each slot, plus a set of general features (e.g. the communication function used by the user, the database search method...) \cite{Henderson2014a}. The set $\mathcal{A}$ is defined as a set of \textit{summary actions}, where the actions can be either slot dependent (e.g. \textit{request}(food), \textit{confirm}(area)...) or slot independent\footnote{We include the summary actions dependent on all the slots, such as \textit{inform}(), in this group.} (e.g. \textit{hello}(), \textit{inform}()...).

The belief space $\mathcal{B}$ is defined by the ontology, therefore belief states of different domains will have different shapes. In order to transfer knowledge between domains, Domain Independent Parametrization (DIP) \citep{wang2015learning} proposes to abstract the belief state $b$ into a fixed size representation. As each action is either slot independent or dependent on a slot $s$, a feature function $\phi_{dip}(b,s)$ can be defined, where $s\in\mathcal{S}\cup s_i$ and $s_i$ stands for slot independent actions. Therefore, in order to compute the policy, $Q(b,a)$ can be approximated as $Q(\phi_{dip}(b,s),a)$, where $s$ is the slot associated to action $a$.

\citet{wang2015learning} presents a handcrafted feature function $\phi_{dip}(b,s)$. It includes the slot independent features of the belief state, a summarised representation of the joint belief state, and a summarised representation of the belief state of the slot $s$. Section \ref{sec:exp_set} gives a more detailed description of the  $\phi_{dip}(b,s)$ function used in this work.
\section{Feudal dialogue management}\label{sec:feu_dm}
\begin{figure}[!t]
  \centering
  \includegraphics[width=\columnwidth,trim={5.8cm 4.8cm 10cm 2.8cm},clip]{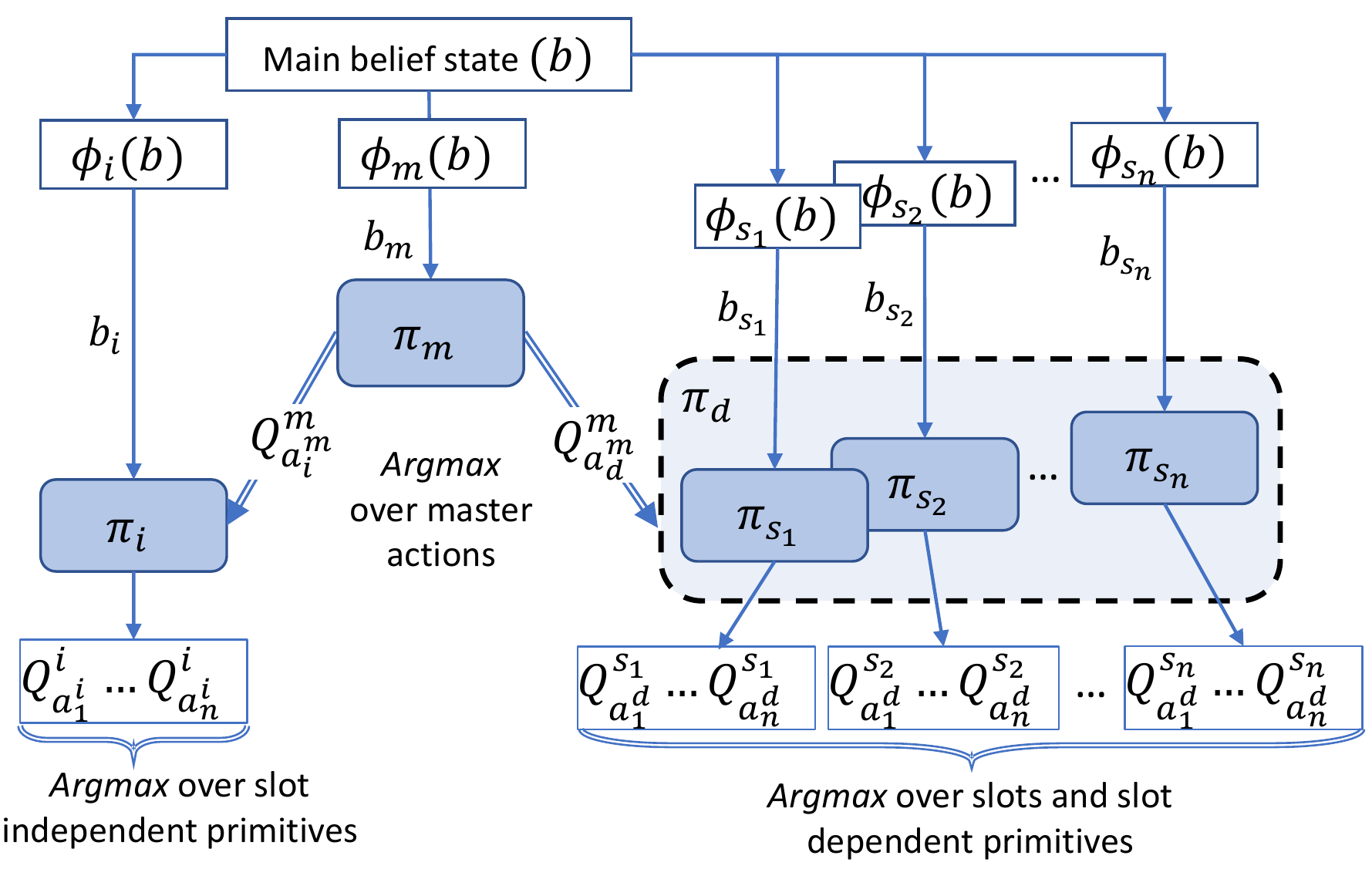}
  \caption{Feudal dialogue architecture used in this work. The sub-policies surrounded by the dashed line have shared parameters. The simple lines show the data flow and the double lines the sub-policy decisions.}
  \label{fig:feudal_arch}
\end{figure}

FRL decomposes the policy decision $\pi(b)=a$ in each turn into several sub-decisions, using different abstracted parts of the belief state in each sub-decision. The objective of a task oriented SDS is to fulfill the users goal, but as the goal is not observable for the SDS, the SDS needs to gather enough information to correctly fulfill it. Therefore, in each turn, the DM can decompose its decision in two steps: first, decide between taking an action in order to gather information about the user goal (information gathering actions) or taking an action to fulfill the user goal or a part of it (information providing actions) and second, select a (primitive) action to execute from the previously selected subset. 
In a slot-filling dialogue, the set of information gathering actions can be defined as the set of slot dependent actions, while the set of information providing actions can be defined as the remaining actions. 

The architecture of the feudal policy proposed by this work is represented schematically in Figure \ref{fig:feudal_arch}. The (primitive) actions are divided between two subsets; slot independent actions $\mathcal{A}_{i}$ (e.g. hello(), inform()); and slot dependent actions $\mathcal{A}_{d}$ (e.g. request(), confirm())\footnote{Note that the actions of this set are composed just by the communication function of the slot dependent actions, thus reducing the number of actions compared to $\mathcal{A}$.}. In addition, a set of master actions $\mathcal{A}_m=(a^m_i, a^m_d)$ is defined, where $a^m_i$ corresponds to taking an action from $\mathcal{A}_{i}$ and $a^m_d$ to taking an action from $\mathcal{A}_{d}$.
Then, a feature function $\phi_s(b)=b_s$ is defined for each slot $s\in\mathcal{S}$, as well as a slot independent feature function $\phi_{i}(b)=b_i$ and a master feature function $\phi_{m}(b)=b_m$. These feature functions can be handcrafted (e.g. the DIP feature function introduced in section \ref{sec:bgnd}) or any function approximator can be used (e.g. neural networks trained jointly with the policy). 

Finally, a master policy $\pi_m(b_m)=a^m$, a slot independent policy $\pi_i(b_i)=a^{i}$ and a set of slot specific policies $\pi_s(b_s)=a^{d}$, one for each $s\in\mathcal{S}$, are defined, where $a^{m} \in \mathcal{A}_{m}$, $a^{i} \in \mathcal{A}_{i}$ and $a^{d} \in \mathcal{A}_{d}$. Contrary to other feudal policies, the slot specific sub-policies have shared parameters, in order to generalise between slots (following the idea used by \citet{Henderson2014b} for DST). The differences between the slots (size, value distribution...) are accounted by the feature function $\phi_s(b)$. Therefore $\pi_m(b_m)$ is defined as:
\begin{equation}
\pi_m(b_m)=\underset{a^m\in\mathcal{A}_m}\argmax Q^m(b_m,a^m)
\end{equation}
If $\pi_m(b_m)=a^m_i$, the sub-policy run is $\pi_{i}$:
\begin{equation}
\pi_{i}(b_i)=\underset{a^i\in\mathcal{A}_i}\argmax Q^i(b_i,a^i)
\end{equation}
Else, if $\pi_m(b_m)=a^m_d$, $\pi_{d}$ is selected. This policy runs each slot specific policy, $\pi_{s}$, for all $s\in\mathcal{S}$, choosing the action-slot pair that maximises the Q function over all the slot sub-policies.
\begin{equation}
\pi_{d}(b_{s}| \forall s \in \mathcal{S})=\underset{a^d\in\mathcal{A}_d, s\in\mathcal{S}}\argmax Q^s(b_s,a^d)
\end{equation}
Then, the summary action $a$ is constructed by joining $a^{d}$ and $s$ (e.g. if $a^d$=\textit{request()} and $s$=\textit{food}, then the summary action will be \textit{request(food)}). A pseudo-code of the Feudal Dialogue Policy algorithm is given in Appendix \ref{apx:feu_alg}. 
\section{Experimental setup}\label{sec:exp_set}
The models used in the experiments have been implemented using the PyDial toolkit \citep{ultes2017pydial}\footnote{The implementation of the models can be obtained in \url{www.pydial.org}} and evaluated on the PyDial benchmarking environment 
 \citep{casanueva2017benchmarking}. This environment presents a set of tasks which span different size domains, different Semantic Error Rates (SER), and different configurations of action masks and user model parameters (Standard (Std.) or Unfriendly (Unf.)). Table \ref{tab:envs} shows a summarised description of the tasks. The models developed in this paper are compared to the state-of-the-art RL algorithms and to the handcrafted policy presented in the benchmarks. 
\begin{table}[t]
\resizebox{0.99\columnwidth}{!}{
\centering
\begin{tabular}{l| c c c c}
Domain & Code & \multicolumn{1}{c}{\# constraint slots} & \multicolumn{1}{c}{\# requests}& \multicolumn{1}{c}{\# values} \\ \hline 
Cambridge Restaurants & CR & 3 & 9 & 268\\ 
San Francisco Restaurants & SFR & 6 & 11 & 636 \\ 
Laptops & LAP & 11 & 21 & 257\\ 
\end{tabular}}
  \resizebox{0.99\columnwidth}{!}{%
    \begin{tabular}{l|cccccc}
          & Env. 1 & Env. 2 & Env. 3 & Env. 4 & Env. 5 & Env. 6 \\ \hline
    SER   & 0\%   & 0\%   & 15\%  & 15\%  & 15\%  & 30\% \\
    Masks & on    & off   & on    & off   & on    & on \\
    User  & Std.     & Std.     & Std.     & Std.     & Unf.     & Std. \\
    \end{tabular}%
    }
  \caption{Sumarised description of the domains and environments used in the experiments. Refer to \citep{casanueva2017benchmarking} for a detailed description.}\label{tab:envs}
\end{table}
\subsection{DIP-DQN baseline}
An implementation of DIP based on Deep-Q Networks (DQN) \cite{mnih2013playing} is implemented as an additional baseline \citep{papangelis2017single}. This policy, named DIP-DQN, uses the same hyperparameters as the DQN implementation released in the PyDial benchmarks. A DIP feature function based in the description in \citep{wang2015learning} is used, $\phi_{dip}(b,s)=\psi_0(b) \oplus \psi_j(b) \oplus \psi_d(b,s) $, where:\\
$\bullet$ $\psi_0(b)$ accounts for general features of the belief state, such as the database search method.\\
$\bullet$ $\psi_j(b)$ accounts for features of the joint belief state, such as the entropy of the joint belief.\\
$\bullet$ $\psi_d(b,s)$ accounts for features of the marginal distribution of slot $s$, such as the entropy of $s$.\\
Appendix \ref{dip_feat} shows a detailed description of the DIP features used in this work.
\subsection{Feudal DQN policy}
A Feudal policy based on the architecture described in sec. \ref{sec:feu_dm} is implemented, named FDQN. Each sub-policy is constructed by a DQN policy \citep{su2017sample}. These policies have the same hyperparameters as the baseline DQN implementation, except for the two hidden layer sizes, which are reduced to 130 and 50 respectively. As feature functions, subsets of the DIP features are used:
\begin{align}
&\phi_m(b)=\phi_i(b)=\psi_0(b)\oplus\psi_j(b) \nonumber\\
&\phi_s(b)=\psi_0(b) \oplus \psi_j(b) \oplus \psi_d(b,s)\, \forall s \in \mathcal{S} \nonumber
\end{align}
The original set of summary actions of the benchmarking environment, $\mathcal{A}$, has a size of $5+3*|\mathcal{S}|$, where $|\mathcal{S}|$ is the number of slots. This set is divided in two subsets\footnote{An additional \textit{pass()} action is added to each subset, which is taken whenever the other sub-policy is executed. This simplifies the training algorithm.}: $\mathcal{A}_{i}$ of size 6 and $\mathcal{A}_{d}$ of size 4. Each sub-policy (including $\pi_m$) is trained with the same sparse reward signal used in the baselines, getting a reward of $20$ if the dialogue is successful or $0$ otherwise, minus the dialogue length. 
\section{Results}\label{sec:res}
The results in the 18 tasks of the benchmarking environment after 4000 training dialogues are presented in Table \ref{tab:res}. The same evaluation procedure of the benchmarks is used, presenting the mean over 10 different random seeds and testing every seed for 500 dialogues.
\begin{table}[t]
  \centering
 
  \resizebox{0.99\columnwidth}{!}{%
    \begin{tabular}{rl||rrrr|rr}
          &       & \multicolumn{2}{c}{Feudal-DQN} & \multicolumn{2}{c}{DIP-DQN} & \multicolumn{1}{|c}{Bnch.} & \multicolumn{1}{c}{Hdc.} \\
    \multicolumn{2}{c||}{\textit{Task}} & \multicolumn{1}{c}{Suc.} & \multicolumn{1}{c}{Rew.} & \multicolumn{1}{c}{Suc.} & \multicolumn{1}{c}{Rew.} & \multicolumn{1}{|c}{Rew.} & \multicolumn{1}{c}{Rew.} \\ \hline \hline
    \multicolumn{1}{r}{\multirow{3}[0]{*}{\rotatebox[origin=c]{90}{Env. 1}}} & CR    & 89.3\% & 11.7  & 48.8\% & -2.8  & 13.5  & \textbf{14.0} \\
          & SFR   & 71.1\% & 7.1   & 25.8\% & -7.4  & 11.7  & \textbf{12.4} \\
          & LAP   & 65.5\% & 5.7   & 26.6\% & -8.8  & 10.5  & \textbf{11.7} \\ \hline
    \multicolumn{1}{r}{\multirow{3}[0]{*}{\rotatebox[origin=c]{90}{Env. 2}}} & CR    & 97.8\% & 13.1  & 85.5\% & 9.6   & 12.2  & \textbf{14.0} \\
          & SFR   & 95.4\% & \textbf{12.4} & 85.7\% & 8.4   & 9.6   & \textbf{12.4} \\
          & LAP   & 94.1\% & \textbf{12.0} & 89.5\% & 9.7   & 7.3   & 11.7 \\ \hline
    \multicolumn{1}{r}{\multirow{3}[0]{*}{\rotatebox[origin=c]{90}{Env. 3}}} & CR    & 92.6\% & 11.7  & 86.1\% & 8.9   & \textbf{11.9} & 11.0 \\
          & SFR   & 90.0\% & \textbf{9.7} & 59.3\% & 0.2   & 8.6   & 9.0 \\
          & LAP   & 89.6\% & \textbf{9.4} & 71.5\% & 3.1   & 6.7   & 8.7 \\ \hline
    \multicolumn{1}{r}{\multirow{3}[0]{*}{\rotatebox[origin=c]{90}{Env. 4}}} & CR    & 91.4\% & \textbf{11.2} & 82.6\% & 8.7   & 10.7  & 11.0 \\
          & SFR   & 90.3\% & \textbf{10.2} & 86.1\% & 9.2   & 7.7   & 9.0 \\
          & LAP   & 88.7\% & \textbf{9.8} & 74.8\% & 6.0   & 5.5   & 8.7 \\ \hline
    \multicolumn{1}{r}{\multirow{3}[0]{*}{\rotatebox[origin=c]{90}{Env. 5}}} & CR    & 96.3\% & \textbf{11.5} & 74.4\% & 2.9   & 10.5  & 9.3 \\
          & SFR   & 88.9\% & \textbf{7.9} & 75.5\% & 3.2   & 4.5   & 6.0 \\
          & LAP   & 78.8\% & 5.2   & 64.4\% & -0.4  & 4.1   & \textbf{5.3} \\ \hline
    \multicolumn{1}{r}{\multirow{3}[0]{*}{\rotatebox[origin=c]{90}{Env. 6}}} & CR    & 90.6\% & \textbf{10.4} & 83.4\% & 8.1   & 10.0  & 9.7 \\
          & SFR   & 83.0\% & \textbf{7.1} & 71.9\% & 3.9   & 3.9   & 6.4 \\
          & LAP   & 78.5\% & \textbf{6.0} & 66.5\% & 2.7   & 3.6   & 5.5 \\
    \end{tabular}%
    }
\vspace*{-0.2cm}
   \caption{Success rate and reward for Feudal-DQN and DIP-DQN in the 18 benchmarking tasks, compared with the reward of the best performing algorithm in each task (Bnch.) and the handcrafted policy (Hdc.) presented in \citep{casanueva2017benchmarking}. 
   }\label{tab:res}
\end{table}%
The FDQN policy substantially outperforms every other other policy in all the environments except Env. 1. The performance increase is more considerable in the two largest domains (SFR and LAP), with gains up to 5 points in accumulated reward in the most challenging environments (e.g. Env. 4 LAP), compared to the best benchmarked RL policies (Bnch.). In addition, FDQN consistently outperforms the handcrafted policy (Hdc.) in environments 2 to 6, which traditional RL methods could not achieve.
In Env. 1, however, the results for FDQN and DIP-DQN are rather low, specially for DIP-DQN. Surprisingly, the results in Env. 2, which only differs from Env. 1 in the absence of action masks (thus, in principle, is a more complex environment), outperform every other algorithm. Analysing the dialogues individually, we could observe that, in this environment, both policies are prone to ``overfit" to an action \footnote{The model overestimates the value of an incorrect action, continuously repeating it until the user runs out of patience.}. The performance of FDQN and DIP-DQN in Env. 4 is also better than in Env. 3, while the difference between these environments also lies in the masks. This suggests that an specific action mask design can be helpful for some algorithms, but can harm the performance of others. This is especially severe in the DIP-DQN case, which shows good performance in some challenging environments, but it is more unstable and prone to overfit than FDQN.

\begin{figure}[t]
  \centering
  \includegraphics[width=\linewidth,trim={0cm 4cm 1.5cm 0cm},clip]{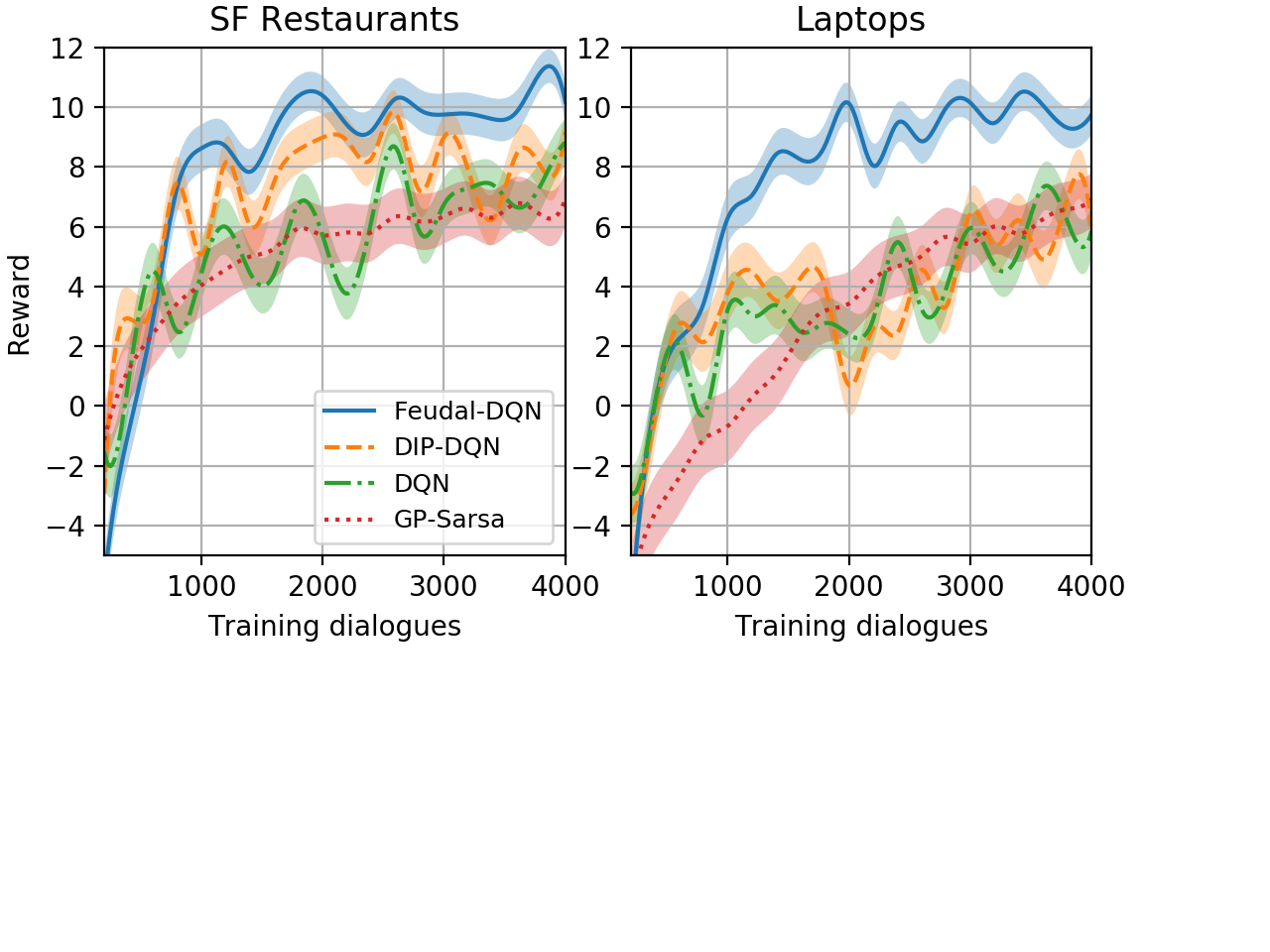}
  \vspace*{-0.5cm}
  \caption{Learning curves for Feudal-DQN and DIP-DQN in Env. 4, compared to the two best performing algorithms in \citep{casanueva2017benchmarking} (DQN and GP-Sarsa). The shaded area depicts the mean $\pm$ the standard deviation over ten random seeds.}
  \label{fig:learn_cur}
\end{figure}
However, the main purpose of action masks is to reduce the number of dialogues needed to train a policy. Observing the learning curves shown in Figure \ref{fig:learn_cur}, the FDQN model can learn a near-optimal policy in large domains in about 1500 dialogues, even if no additional reward is used, making the action masks unnecessary.
\section{Conclusions and future work}
We have presented a novel dialogue management architecture, based on Feudal RL, which substantially outperforms the previous state of the art in several dialogue environments. By defining a set of slot dependent policies with shared parameters, the model is able to learn a general way to act in slots, increasing its scalability to large domains. 

Unlike other HRL methods applied to dialogue, no additional reward signals are needed and the hierarchical structure can be derived from a flat ontology, substantially reducing the design effort.


A promising approach would be to substitute the handcrafted feature functions used in this work by neural feature extractors trained jointly with the policy. This would avoid the need to design the feature functions and could be potentially extended to other modules of the SDS, making text-to-action learning tractable. In addition, a single model can be potentially used in different domains \citep{papangelis2017single}, and different feudal architectures could make larger action spaces tractable (e.g. adding a third sub-policy to deal with actions dependent on 2 slots).

\section*{Acknowledgments}

This research was funded by the EPSRC grant EP/M018946/1 Open Domain Statistical Spoken Dialogue Systems

\bibliography{biblio.bib}
\bibliographystyle{acl_natbib}
\appendix

\section{Feudal Dialogue Policy algorithm}
\label{apx:feu_alg}

\begin{algorithm}
  \caption{Feudal Dialogue Policy}\label{alg:feudal_policy}
  \begin{algorithmic}[1]
    \For{\text{each dialogue turn}}
      \State observe $b$
      \State $b_m=\phi_m(b)$
      \State $a^m=\underset{a^m\in\mathcal{A}_m}\argmax Q^m(b_m,a^m)$ 
      \vspace*{0.2cm}
      \If{$a^m==a^{m}_i$} \Comment{drop to $\pi_{i}$}
      \State $b_i=\phi_i(b)$
      \State $a=\underset{a^i\in\mathcal{A}_i}\argmax Q^i(b_i,a^i)$
      \vspace*{0.2cm}
      \Else{\text{ }$a^m==a^{m}_d$ \text{\textbf{then}}} \Comment{drop to $\pi_{d}$}
      \State $b_s=\phi_s(b)\, \forall s \in \mathcal{S}$
      \State $slot, act = \underset{s\in\mathcal{S},a^{d}\in\mathcal{A}_{d}}\argmax Q^s(b_s,a^{d})$
      \vspace*{0.2cm}
      \State $a=join(slot, act)$
      \EndIf 
      
      \State execute $a$
    \EndFor
  \end{algorithmic}
 \end{algorithm}
\newpage
\section{DIP features}\label{dip_feat}
This section gives a detailed description of the DIP feature functions $\phi_{dip}(b,s)=\psi_0(b) \oplus \psi_j(b) \oplus \psi_d(b,s) $ used in this work. The  differences with the features used in \citep{wang2015learning} and \citep{papangelis2017single} are the following:
\begin{itemize}
\item No \textit{priority} or \textit{importance} features are used. 
\item No \textit{Potential contribution to DB search} features are used. 
\item The joint belief features $\psi_j(b)$ are extended to account for large-domain aspects.
\end{itemize}

\begin{table}[h]
  \centering
  \resizebox{0.99\columnwidth}{!}{%
    \begin{tabular}{rr|l}
    \multicolumn{1}{l|}{Feature} & \multicolumn{1}{l|}{Feature} & Feature \\
    \multicolumn{1}{l|}{function} & \multicolumn{1}{l|}{description} & size \\
    \midrule
    \multicolumn{1}{l|}{$\psi_0(b)$} & \multicolumn{1}{l|}{last user dialogue act (bin) *} & 7 \\
    \multicolumn{1}{r|}{} & \multicolumn{1}{l|}{DB search method (bin) *} & 6 \\
    \multicolumn{1}{r|}{} & \multicolumn{1}{l|}{\# of requested slots (bin)} & 5 \\
    \multicolumn{1}{r|}{} & \multicolumn{1}{l|}{offer happened * } & 1 \\ \multicolumn{1}{r|}{} & \multicolumn{1}{l|}{last action was \textit{Inform no venue} *} & 1 \\
    \multicolumn{1}{r|}{} & \multicolumn{1}{l|}{normalised \# of slots (1/\# of slots)} & 1 \\
    \multicolumn{1}{r|}{} & \multicolumn{1}{l|}{normalised avg. slot length (1/avg. \# of values)} & 1 \\
    \midrule
    \multicolumn{1}{l|}{$\psi_j(b)$} & \multicolumn{1}{l|}{prob. of the top 3 values of $b_j$} & 3 \\
    \multicolumn{1}{r|}{} & \multicolumn{1}{l|}{prob. of *NONE* value of $b_j$} & 1 \\
    \multicolumn{1}{r|}{} & \multicolumn{1}{l|}{entropy of $b_j$} & 1 \\
    \multicolumn{1}{r|}{} & \multicolumn{1}{l|}{diff. between top and 2nd value probs. (bin)} & 5 \\
    \multicolumn{1}{r|}{} & \multicolumn{1}{l|}{\# of slots with top value not *NONE* (bin)} & 5 \\
    \midrule
    \multicolumn{1}{l|}{$\psi_d(b,s)$} & \multicolumn{1}{l|}{prob. of the top 3 values of $s$} & 3 \\
    \multicolumn{1}{r|}{} & \multicolumn{1}{l|}{prob. of *NONE* value of $s$} & 1 \\
    \multicolumn{1}{r|}{} & \multicolumn{1}{l|}{diff. between top and 2nd value probs. (bin)} & 5 \\
    \multicolumn{1}{r|}{} & \multicolumn{1}{l|}{entropy of $s$} & 1 \\
    \multicolumn{1}{r|}{} & \multicolumn{1}{l|}{\# of values of $s$ with prob. $>$ 0 (bin)} & 5 \\
    \multicolumn{1}{r|}{} & \multicolumn{1}{l|}{normalised slot length (1/\# of values)} & 1 \\
    \multicolumn{1}{r|}{} & \multicolumn{1}{l|}{slot length (bin)} & 10 \\
    \multicolumn{1}{r|}{} & \multicolumn{1}{l|}{entropy of the distr. of values  of $s$ in the DB} & 1 \\
\cmidrule{2-3}          & total & 64 \\
    \end{tabular}%
    }
  \caption{List of features composing the DIP features. the tag (bin) denotes that a binary encoding is used for this feature. Some of the joint features $\psi_j(b)$ are extracted from the joint belief $b_j$, computed as the Cartesian product of the beliefs of the individual slots. * denotes that these features exist in the original belief state $b$.}
  \label{tab:dip_feats}%
\end{table}%

\end{document}